\documentclass{article}

\usepackage{arxiv}

\usepackage[utf8]{inputenc} 
\usepackage[T1]{fontenc}    
\usepackage{hyperref}       
\usepackage{url}            
\usepackage{booktabs}       
\usepackage{amsfonts}       
\usepackage{nicefrac}       
\usepackage{microtype}      
\usepackage{lipsum}
\usepackage{graphicx}
\usepackage{algorithm}
\title{Handling Imbalanced Data: A Case Study for Binary Class Problems}

\author{
  Richmond Addo~Danquah \\
  Department of Mathematics\\
  Southern Illinois University\\
  Edwardsville, IL 62026 \\
  \texttt{raddoda@siue.edu} \\
}

\begin{document}
\maketitle

\begin{abstract}

   For several years till date, the major issues in terms of solving for classification problems are the issues of Imbalanced data. Because majority of the machine learning algorithms by default assumes all data are balanced, the algorithms do not take into consideration the distribution of the data sample class. The results tend to be unsatisfactory and skewed towards the majority sample class distribution. This implies that the consequences as a result of using a model built using an Imbalanced data without handling for the Imbalance in the data could be misleading both in practice and theory. Most researchers have focused on the application of Synthetic Minority Oversampling Technique (SMOTE) and Adaptive Synthetic (ADASYN) Sampling Approach in handling data Imbalance independently in their works and have failed to better explain the algorithms behind these techniques with computed examples. This paper focuses on both synthetic oversampling techniques and manually computes synthetic data points to enhance easy comprehension of the algorithms. We analyze the application of these synthetic oversampling techniques on binary classification problems with different Imbalanced ratios and sample sizes.
\end{abstract}


\section{Introduction}
\begin{figure}
  \centering
  \includegraphics[width=10cm]{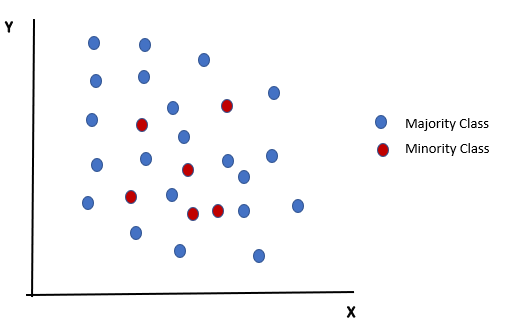}
  \caption{A graphical representation of an Imbalanced Binary Data.}
  \label{Schematic diagram of an Imbalanced Binary Data}
\end{figure}

When it comes to solving classification problems using machine learning algorithms, the issue of Imbalanced data has become pervasive and ubiquitous. As shown in Figure \ref{Schematic diagram of an Imbalanced Binary Data}, a data is said to be Imbalanced if the sample of one class is in a higher number than the other class. The class with the highest number in the data is referred to as the majority sample class and the class with the least number in the data is minority sample class. While we may assume that machine learning classifiers should be able to classify any data without any bias towards a certain sample class, this is not the case when the machine learning classifier is fed with an Imbalanced data. By default, these classifiers assume that there are even sample class distributions in the data and so tend to produce unsatisfactory results when the sample class distributions are Imbalanced. These results tend to be biased toward the majority class distribution. One could understand why the significance of this area of research continues to grow because in most cases, the focus of interest lies in the minority sample class. And if the machine learning is favoring the majority sample class, that implies we are getting the opposite result of what we expect to get. In most cases and especially in real world applications, Imbalanced data are born 'naturally". That means they are inherently in the data. Examples of the naturally inherent Imbalanced data include but not limited to people with a certain chronic condition vs. those without, people who commit bank fraud vs those who do not, employees who churn vs those who stay. The effect of modeling or predicting using an Imbalanced data can cause a greater damage than good. Take for example, in a financial institution where there is critical need to identify fraud transaction which are mostly rare. Any errors in detecting a fraudulent transaction causes a major financial blow to the company. The same applies to the medical diagnosis field where certain health conditions are rare and therefore physicians or doctors cannot afford to incorrectly diagnose a patient. The effect of such incorrect diagnosis could be extremely dangerous to the patient.\par
Handling Imbalanced classification problems can be approached in two ways - either through allocating different costs to training examples or re-sample the original data, either by over-sampling the minority class or by under-sampling the majority class \cite{chawla2002SMOTE}. In this paper, the focus will be on re-sampling of the original data set using synthetic oversampling techniques, SMOTE (Synthetic Minority Oversampling Technique), and its extension ADASYN (Adaptive Synthetic) technique. Section 2 gives an overview of reporting measures; Accuracy, Recall, Precision, F1-Score and Area Under the  Receiver Operator Characteristic (ROC) Curve (AUC). Section 3 presents the details of our oversampling techniques, SMOTE and ADASYN. Section 4 presents data analysis results comparing the two oversampling techniques to each other using some machine learning classifiers. Section 5 concludes the paper and offer recommendations for future work.

\section{Reporting Measures}
\begin{figure}[!bht]
\centering
\includegraphics[width=8cm]{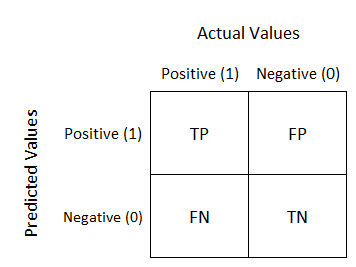}
\caption{The Confusion Matrix }
\label{Confusion Matrix}
\end{figure}

In evaluating the performance of a classifier, we use the confusion matrix. A confusion matrix has 4 quadrants and each of the observation data points in a classification problem  belongs to one of the possible four outcomes as shown in Figure \ref{Confusion Matrix}. As shown in Figure \ref{Confusion Matrix}, the True Positive (TP) quadrant shows the samples that are positive and have been classified as Positive, the False Positive (FP) quadrant shows the samples that are negative but have been classified as positive, the True Negative (TN) quadrant shows samples that are negative and have been classified as negative and the False Negative (FN) quadrant shows samples that are positive but have classified as negative.  A simple computation involving two or more of the quadrant: True Positive , False Positive, True Negative and False Negative results in the models reporting measures. These include but not limited to Accuracy, Precision, Recall/Sensitivity, F1-score, AUC score, G-mean, Discriminant Power among other reporting measures. Since this study focused on binary-class problems, we  consider model's accuracy, precision, sensitivity/recall, F1-score and Area Under the Curve (AUC) score as our performance measures on the models.
\newpage
It is also important to know how the reporting measures are calculated and most importantly what they mean. 
\begin{itemize}
   \item \textbf{Accuracy} is the most popular evaluation metric in the data community. But accuracy is a great reporting measure  when there is a balanced data set between the sample classes. And because most classification problems especially real world problems are naturally Imbalanced, relying on this measure could be misleading to solving the problem. It is therefore recommended that we consider other measurements when evaluating model performance. Accuracy can be calculated mathematically as: 
   \begin{equation}
    Accuracy=\frac{TP + TN}{TP+ TN + FP + FN} 
   \end{equation}

\item \textbf{Precision} is the ratio between True Positives observations and all Positive observations. That is, of all the instances the classifier labeled as positive, what fraction were actually correct? Also called Positive Predictive Value (PPV), it can be described as a measure of a classifiers exactness. Mathematically,
\begin{equation}
   Precision = \frac{TP}{TP + FP}  
\end{equation}

\item \textbf{Recall/ Sensitivity} is a measure of a classifiers completeness. Recall can be described as the ration of the number of positive predictions and the number of positive class values, that is, the ability of the classifier to find all positive instances. Mathematically,
\begin{equation}
 Recall = \frac{TP}{TP + FN}
 \end{equation}
 
\item \textbf{F1-score} is normally considered as the model's accuracy that takes into consideration Precision and Recall. And this is particularly very useful and not misleading compared to accuracy especially when dealing with asymmetric or imbalance data set. F1-Score takes into account the weighted harmonic mean of both the classifiers exactness (Precision) and completeness (Recall/Sensitivity). The higher the Precision and Recall, the higher F1-score will be. This measure optimizes Precision and Recall into a single value. Mathematically,
\begin{equation}
F1-Score = \frac{2*(Recall * Precision)}{Recall + Precision} 
\end{equation}
\item \textbf{AUC (Area Under the Receiver Operator Characteristic (ROC) Curve) Score}
 By plotting the True Positive Rate against the False Positive Rate at various threshold levels, we create the ROC graph. AUC is simply the area under the ROC curve which shows the  likelihood that a classifier would rank a positive observation randomly chosen higher than a negative observation randomly chosen. It is best understood to be a measure of separability - the ability of the classifier to distinguish between the sample class distribution. The closer the AUC score is to 1, the better the classifier is able to distinguish between the minority and majority sample class. When faced with an Imbalanced data, this is a better measure to report than accuracy. AUC score measure can be computed as;
 \begin{equation}
 AUC = \frac{1 + TP - FP}{2} 
\end{equation}
\end{itemize}

\section{Oversampling Technique}
For many years, there has been an extended research on the use of oversampling techniques in solving class imbalance problems. The extended research on this technique can be attributed to its ability to retain the original data set while preventing the loss of important information. Some of these researches on oversampling technique were developed by notable researchers such as (Douzas \& Bação, 2018), (Last, Douzas, \& Bacao, 2017), (Nekooeimhr \& Lai-Yuen, 2016), (Li, Fong, Wong, Mohagmmed, \& Fiaidhi , 2016), (Sun, Song , Zhu, Xu, \& Zhou, 2015), (Menardi \& Torelli, 2014) and (Bowyer, Hall, Kegelmeyer, \& Chawla, 2002). This technique creates a balanced data set by generating new samples to be added to the minority sample class. Oversampling can be done either through random oversampling where the data set is balanced through replicating the existing minority sample class or through synthetic oversampling where the data set is balanced through creating new synthetic minority samples by linear interpolation. The focus of this study will be on the application of synthetic oversampling in handling Imbalanced data because this method unlike random oversampling avoids over fitting, therefore improving the generalization ability of the classifier.

\subsection{3a. Synthetic Minority Oversampling Technique (SMOTE)}
This approach, inspired by a technique that proved successful in handwritten character recognition (Ha \& Bunke, 1997),  Synthetic Minority Oversampling Technique (SMOTE) was first pioneered by Chawla in 2002 \cite{chawla2002SMOTE}. In SMOTE algorithm, minority class is over sampled by generating synthetic examples rather than by oversampling with replacement for simple random oversampling. To avoid the issue of over fitting when increasing minority sample class, SMOTE creates synthetic data points by working within the current feature space. New synthetic data points  are extracted from interpolation, so the original data set still has significance. SMOTE interpolates values using a K - nearest neighboring technique for each minority class instance and generates attribute values for new data instances \cite{Maheshwari2011approach}.\par A new synthetic data point is created for each minority sample data by taking the difference between the a minority sample class feature vector and the nearest neighbor belonging to the same sample class and multiplying it by a random number between 0 and 1 and then adding the results back to the minority sample class feature vector. This creates a random line segment between every pair of existing features. This results in a new instance generated  within the data set \cite{deepa2013textural}. The cycle is replicated for the remaining minority sample data \cite{Satyasree2013literature}.\par
One such downside stems from the fact that SMOTE arbitrarily tries to over-sample a minority instance with a uniform likelihood. Although this helps the approach to efficiently counter imbalances between the classes, the problems of disparity within the class and small disjoints are overlooked. Input areas that report multiple minority populations are highly likely to be further inflated, while sparsely populated minority areas are likely to remain sparse (Prati et al., 2004). This increases the complexity of the problem and lowers the learning classifier efficiency. Another downside is that SMOTE will intensify the noise present in the data further. This is likely to happen when a noisy minority sample, which is situated between instances of the majority class, and its nearest minority neighbor interpolates linearly. The approach is susceptible to noise generation, as it does not differentiate between overlapping class regions and so-called protected areas (Bunkhumpornpat et al., 2009). This impedes the classifier's ability to define the problems boundaries \cite{Last2017oversampling}.

\begin{figure}[bht]
\centering
\includegraphics[width=15cm]{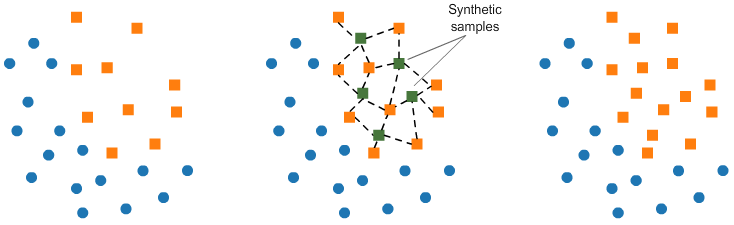}
\caption{A Schematic Diagram on how to create a Synthetic data using the SMOTE Technique. }
\label{Schematic diagram of SMOTE  }
\end{figure}
An Imbalanced class distribution (LEFT) as shown in Figure \ref{Schematic diagram of SMOTE  } contains more blue colors than orange colors. Using SMOTE, the algorithm finds the K-nearest neighbour of a data point in the minority class (orange colors) and creates some synthetic data points on the lines joining the primary point and the neighbors as shown in Figure \ref{Schematic diagram of SMOTE } (middle). These new neighbors synthetic data points generated share similar characteristics of the other minority data points. Theses synthetic data points now help balance the original class distribution (right), which improves the model's generalization ability. The use of SMOTE for class distribution balancing ensures no loss of data, alleviates over fitting and the process is very easy to implement. However, SMOTE should be used with extra care when dealing with higher dimensional space. \par
Researchers and practitioners have widely embraced SMOTE, perhaps due to its versatility and added benefit with regard to random oversampling \cite{Last2017oversampling}. The table below has been adopted and modified from the original research by (Chawla et al ...2002) to simplify the processes of the SMOTE algorithm.

\begin{algorithm}

Input:  \\
Let {$x_1$, $x_2$,...,$x_n$} be the minority class feature vectors in the n dimensional space of X \\
Let N be the number of synthetic instances to generate \\
Let K be the number of nearest neighbour \\
Output: \\
Synthetic set of artificial instances
\\
1. For i in range (N) do,\\
        2. Select randomly a minority class feature vector $x_i$\\
        3. From $x_i$'s K-nearest minority class neighbors, randomly select a neighbor $\hat x_{i}$\\
       4.  diff = $\hat x_{i}$ - $x_i$\\
        $\delta = $random number between 0 and 1$\\
       5.  $new Sample$ = $ $x_i$ + diff * $\delta   $ \\
       6.  $Synthetic \Longleftarrow new Sample$  \\
        endfor
\label{A Pseudo Code for SMOTE Algorithm}
\caption{SMOTE}
\end{algorithm}  \par

 To better appreciate how SMOTE works, this paper created an Imbalanced data with 10 data points - three(3) minority class and seven (7) majority class as shown in  Table 1. We then used this Imbalance data to create two (2) minority synthetic data points by using the SMOTE algorithm steps as shown in the table above. For the purpose of illustration, we set K (number of nearest neighbors) to be 2.
 
  \begin{table}[ht]
\caption{Example of Imbalanced Data} 
\centering 
 \begin{tabular}{|c c c c c c c c c c|} 
 \hline\hline
 No & No & No & Yes& Yes & Yes & Yes & Yes & Yes & Yes \\
 \hline
 5 & 4 &5 & 2&1&3&4&4&5&5 \\
 3 & 3 & 2 &6&4&2.5&3&4.5&5&6 \\
 \hline
\end{tabular}
\end{table}

 \begin{algorithm}
Input:  \\
Let N = 2 \\
Let K = 2 \\
Output: \\
Synthetic  set of artificial instances
\\
1. For i in range (N=1) do,\\
        2. Select (4 3)\\
        3. Randomly select a neighbor (5 3)\\
       4.  diff = (5 3) - (4 3) = (1 0) \\
       $\delta = 0.5\\
       5.  $new Sample$ =  $(4 3) + [(1 0) * 0.5] $ \\
       6.  Synthetic $1$ \Longleftarrow $(4.5 3)$ \\
       \\~\\
        $1. for i in range (N=2) do,$\\
      2. $Select (5 2)$\\
      3. $Randomly select a neighbor (5 3)$\\
       4.  $diff = (5 3) - (5 2) = (0 1)$ \\
       5.  $new Sample$ =  $(5 2) + [(0 1) * 0.5] $ \\
       6.  Synthetic $2$ \Longleftarrow $(5 2.5) \\
        end for \\
\caption{Example for SMOTE}
\end{algorithm} \par

 As shown in Figure \ref{SMOTE Application } below, we have manually calculated 2 synthetic data points (4.5 3) and (5 2.5) by using the steps in the SMOTE algorithm. This increased our minority class from three (3) to five (5), therefore improving the imbalance ratio significantly. We then plot the new synthetic data, (4.5 3) and (5 2.5) together with the original data set as shown in Figure \ref{SMOTE Application }. 
 
\begin{figure}[bht]
\centering
\includegraphics[width=16cm]{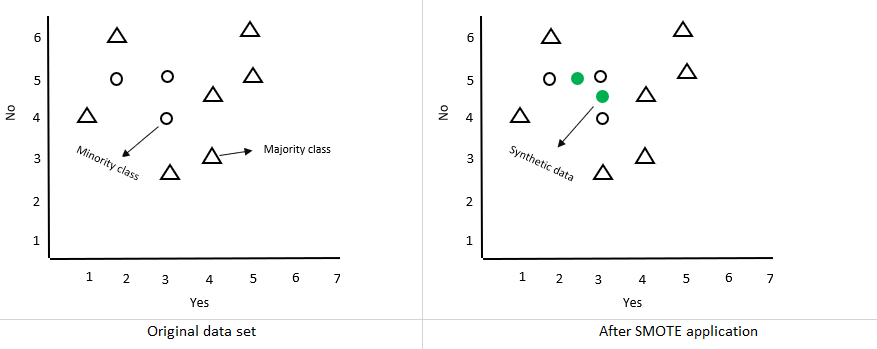}
\caption{A schematic diagram of the Binary Data before and after the application of SMOTE algorithm.} 
\label{SMOTE Application }
\end{figure} 

\newpage
\subsection{3b. Adaptive Synthetic (ADASYN) Sampling Approach}
Adaptive Synthetic sampling approach is an extension or improvement of the SMOTE algorithm. The concept of Adaptive Synthetic (ADASYN) sampling approach for
imbalanced learning  was first introduced by He
(2008). ADASYN tries to generate more synthetic instances
on the region with less positive instances than one with more
positive instances to increase the recognition of positive.
This algorithm uses the number of negative neighbors in K-nearest neighbors of each positive instance to form a distribution function. The distribution function determines how many synthetic instances are generated from that positive instance \cite{Wacha2017Adaptive}. For harder to learn minority samples, ADASYN generates more synthetic data points/observations as compared to minority sample class that are easier to learn. Ultimately, ADASYN is a pseudo-probabilistic algorithm in the sense that a fixed number instances is generated for each minority instance based on a weighted distribution of its neighbors \cite{Haibo2008ADASYN}. Using SMOTE, each minority sample class has equal chance to be selected in the creation of the  synthetic data samples but ADASYN uses the density distribution as a criterion to automatically evaluate the number of synthetic data samples to be produced for each example of minority sample class data \cite{HuangClassification2015}.\par 
Therefore, ADASYN approach improves data distribution learning by reducing the bias generated by the class disparity and moving the classification decision boundary to the challenging examples (He et al, 2008). Because ADASYN is very sensitive to outliers, it is advisable to deal with outliers during data preprocessing before applying ADASYN procedure. Compared to SMOTE, ADASYN put more focus on the minority samples that are difficult to learn by generating more synthetic data points for these difficult and hard to learn minority class samples \cite{Last2017oversampling}.

\begin{figure}[bht]
\centering
\includegraphics[width=12cm]{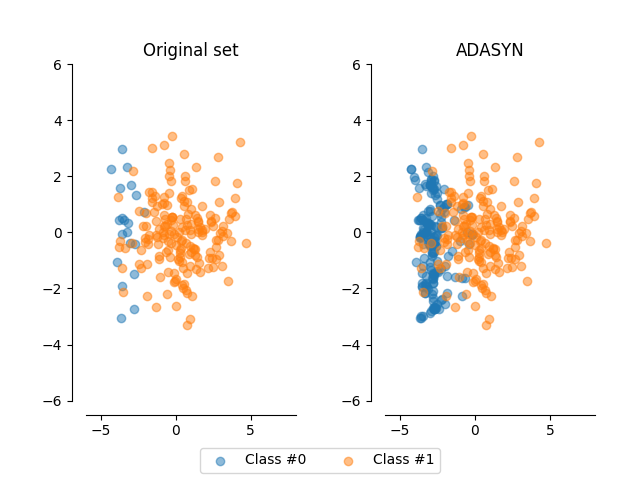}
\caption{Schematic diagram of ADASYN algorithm application. }
\label{Schematic diagram of ADASYN }
\end{figure} 
 As shown in Figure \ref{Schematic diagram of ADASYN }, the original data set has some minority class examples that are difficult for an algorithm to learn. ADASYN synthetically generates new minority samples using a density distribution based on the number of majority sample class neighbors. A minority sample class surrounded with more nearest neighbors from the majority sample class is considered hard-to-train. Therefore the algorithm will give priority to the hard-to-learn minority sample class by generating more synthetic points/observations than the less hard-to-learn minority sample class. The table below shows the pseudo algorithm for ADASYN and has been adopted from (He et al 2008) and modified. \par
\begin{algorithm}
Input:  \\
Let m be the number of minority samples \\
Let n be the number of majority samples\\
Let $\beta $ be the ratio of the balance level of the synthetic samples. NB: $ \beta $$\in$ $(0,1]$ \\
Let $x_i$ for i=1,2,3...m be the minority class feature vectors in the n dimensional space of X \\
Let G be the number of synthetic instances to generate \\
Let $g_i$ for i=1,2,3...m be the number of synthetic data generated for each $x_i$\\
Let K be the Number of Nearest Neighbour \\
Let  $\delta $$\in$ [0,1] \\
Output: \\
Synthetic set of artificial instances \\
$G = \beta \times (n-m) $\\
For i in range (m),\\
Find K for every $x_i$\\
 Calculate    $r_i$ = $f_k$/K , where  $f_k$   is the number of feature vectors in the K nearest neighbors belonging to the majority class  \\  
 Calculate    $\hat r_{i}$ =   $r_i$ / $\sum_{i=1}^{m} r_{i}$, so that $(\sum_{i=1} \hat r_{i} = 1)$ \\
Calculate   $g_i$ =  $\hat r_{i} \times G $\\
For i in range ($g_i$) and for j in range ($m$), do \\
From $x_i$'s K-nearest minority class neighbors, randomly select a neighbor $\hat x_{i_j}$ \\
diff = $\hat x_{i_j}$ - $x_i$\\
$New Sample_{i_j}$ = $x_i$ + diff * $\delta \\
$Synthetic $\Longleftarrow  New Sample_{i_j}$ \\
\label{A Pseudo-code for ADASYN Algorithm}
\caption{ADASYN}
\end{algorithm}

We computed new synthetic minority data points using the ADASYN algorithm using the same data points used in SMOTE application. We created three (3) minority synthetic data points by using the ADASYN algorithm steps as shown in the table above. For the purpose of illustration, we set K to be 2.

\begin{algorithm}

Input:  \\
Let m = 3, n = 7, K = 2, $\beta $ = 0.75 \&  $\delta$ = 0.5\\
Output: \\
1. $G = (7-3) * 0.75 = 3 $\\
2.   ** There is only 1 majority class in each of the distinct neighbourhood \\
i.     $r_i$ = $1/2$ for $i=1$ to $3$ \\
ii. $\sum_{i=1}^{3} r_{i}$=1/2 + 1/2 + 1/2 = 3/2\\
3. $\hat r_{i}$ = 1/2 * 2/3 = 1/3 for $i=1$ to $3$\\
4. $g_i$ = $1/3 * 3 = 1$ for $i=1$ to $3$ \\
     5. From (4 3) 2-nearest minority class neighbors, randomly select a neighbor (5 3)\\
      i.  diff = (5 3) - (4 3)\\
    ii.$New Sample_{i_j}$ = (4 3) + [(1 0) * 0.5] \\
       iii. Synthetic $\Longleftarrow  $(4.5 3)$\\
       \\
      $ 6. From (5 3) 2-nearest minority class neighbors, randomly select a neighbor (5 2)$\\
      i.  $diff$ = $(5 2)$ - $(5 3)$\\
    ii.$$New Sample_{i_j}$ = (5 3) + [(0 -1) * 0.5] \\
       iii. Synthetic $\Longleftarrow  $(5 2.5)$\\
       \\
       $ 7. From (5 2) 2-nearest minority class neighbors, randomly select a neighbor (4 3)$\\
      i.  $diff$ = $(4 3)$ - $(5 2)$ \\
    ii.$$New Sample_{i_j}$ = (5 2)+ [(-1 1) * 0.5] \\
       iii. Synthetic $\Longleftarrow  $(4.5 2.5)\\
        end for
        
\caption{Example for ADASYN}
\end{algorithm}

\begin{figure}[bht]
\centering
\includegraphics[width=16cm]{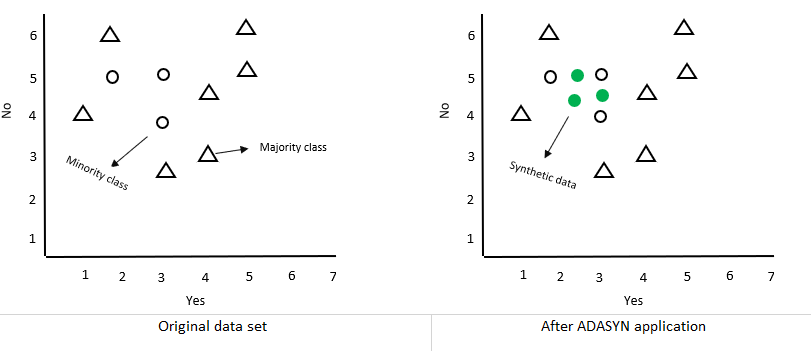}
\caption{A schematic diagram of the Binary Data before and after the application of ADASYN algorithm.}
\label{ADASYN Application }
\end{figure}

 We manually calculated 3 synthetic data points (4.5 3), (5 2.5) and (4.5 2.5) by using the steps in the ADASYN algorithm and data points from Table 1. This increased our minority class from three (3) to six (6), therefore improving the Imbalance ratio significantly. We plot the new synthetic data points,  (4.5 3), (5 2.5) and (4.5 2.5) together with the original data set as shown in Figure \ref{ADASYN Application }. 

\newpage
\section{Data Set Summary \& Pre-Processing}

Using real world data sets obtained from Kaggle and UCI Machine Learning Repository , we test learning capabilities of the SMOTE and ADASYN algorithm on two-class imbalance data. Table 2 shows a brief description of the data sets used in this paper.

\begin{table}[ht]
\caption{Description of data sets} 
\centering 
\begin{tabular}{|c|c|c||c|c|c|} 
\hline 
Data Set & Attributes & Sample Size & Minority & Majority  \
\\ [0.1ex] 
\hline 
Blood Transfusion Service Center& 5 & 748 & 178 & 570  \\
Pima Indians Diabetes &8 &768 & 268 & 500  \\ 
IBM HR Analytics Employee Attrition & 35& 1470 & 237& 1233 \\
 [1ex] 
\hline 
\end{tabular}
\end{table}

\subsection{Blood Transfusion Service Center Data}
This data set originates from the Blood Transfusion Service Center in Hsin-Chu City in Taiwan. The objective of this data set is to use certain measurements to predict whether a person will donate blood. The data set has 5 variables or features with 748 data examples. There are 178 minority class and 570 majority class. This data set is a two-class Imbalanced data set and hence no modification was needed.

\subsection{Pima Diabetes Data}
This data set is owned by the National Institute of Diabetes and Digestive and Kidney Diseases. There are 768 data examples with 8 attributes. This data set is a two class Imbalanced data set with a binary outcome, 0 (not diabetic) and 1 (diabetic) with the aim to predict using certain measurements whether a person is diabetic or not.

\subsection{IBM HR Analytics Employee Attrition Data}
This fictitious data set was created by IBM data scientists for the purpose of uncovering the factors that lead to employee attrition. There are 1470 data sample with 237 minority sample class and 1233 majority sample class. \par

All of the three different data set were checked for missing values, fortunately there were none for all of the data sets. Certain features were converted from categorical values to numerical values to boost performance measures. The outcome variables for IBM data set were re-coded from a nominal binary outcome to a numerical binary outcome. For the purpose of been able to interpret the output and to avoid information loss, we did not perform any dimensional reduction technique especially to the IBM HR Analytics Employee Attrition data set. Again, some of the classifiers used in this paper inherently deal with dimension reduction while learning the data. All of our data sets were standardized to provide a common ground for all the classifiers even though some classifiers can still perform well whether the data is standardized or not. The data sets were split into 80\% train data and 20\% test data. The classifiers first learned using the train data and the test data was used to measure the classifiers learning abilities.\par  
This paper made use of 4 classification classifiers to test the learning abilities of the algorithms on all of the data sets. These classifiers included Logistic Regression, Support Vector Machine (SVM), Random Forest and XGBoost.

\section{Experimental Results}
Combined with SMOTE or ADASYN, 4 classifiers; Logistic Regression, Support Vector Machine (SVM), Random Forest and XGBoost were used as the learning models in our experiment. Without applying SMOTE or ADASYN algorithms, we created baseline models using the original data sets and provided the reporting measures using each of the classifiers mentioned. The reported measures were used to illustrate the performances of these learning techniques and we compared the learning ability of the classifiers before and after performing SMOTE and ADASYN. The following tables below contain the experimental results. The best classifier for each of the three applications ( baseline, SMOTE and ADASYN) and the optimal reported measures in terms of F1-Score and AUC Score are highlighted with bold text. F1-Score and the AUC Score are given much importance because these measures are less leading and provide a more accurate evaluation of the model performance especially when the data is Imbalanced compared to accuracy.\par
The baseline models for all of the three data sets reported high accuracy greater than 70\% when applied to any of the 4 classifiers used in this paper. Although the baseline models reported high accuracy, the harmonic mean which is the F1-Score were relatively low. The reported F1-Score for the Pima Diabetes Data (Table \ref{PIMA}) were much higher than the same score for the Blood Transfusion Service Center Data (Table \ref{BTSC}) and the IBM HR Analytics Employee Attrition Data (Table \ref{IBM}) for the baseline models. Again, the AUC Score compared to to accuracy were also much lower for the baseline models. AUC Score for the Pima Diabetes Data were higher than the other two data sets. And this could be as a result of the differences in data Imbalance among other issues such as sample size, and type of classifier used. While the minority sample class in the Pima Diabetes Data accounted for 35\% of the data, IBM HR Analytics Attrition Data and the Blood Transfusion Service Center Data minority sample class accounted for 24\% and 16\% respectively in the data.\par
There were significant improvement in model performances after handling for the data Imbalance through the application of SMOTE and ADASYN. As shown in Table \ref{BTSC}, \ref{PIMA}, \ref{IBM}, all of the reported measures showed higher scores compared with the base models for each of the classifier combination. The improvement in the scores confirmed that indeed machine learning classifiers are less sensitive to sample class distribution in a data. Most importantly , the ability of the classifiers to distinguish between majority and minority sample class was improved in all of the data sets after the application of SMOTE and ADASYN

\begin{table}[!ht]
\caption{Blood Transfusion Service Center Data }
\label{BTSC}
\begin{tabular}{ |p{3cm}||p{2cm}|p{2cm}|p{2cm}|p{2cm}|p{2cm}|}
 \hline
 \multicolumn{6}{|c|}{\textbf{Original Data Set}} \\
 \hline
 Classifier & Accuracy&Precision&Recall&F1-score&AUC\\
 \hline
 Logistic Regression  &{0.73}    &0.57& 0.10&0.20& 0.54\\
 SVM&   0.74  &0.67 & 0.10   &0.20&0.54\\
 Random Forest&   0.71 &0.45 & 0.32 &0.37 & 0.59\\
 \textbf{XGBoost}& 0.75 &0.58&0.37&{0.45}&{0.63}\\
 \hline
 \multicolumn{6}{|c|}{\textbf{SMOTE}} \\
 \hline
 Logistic Regression   &0.75    &0.74& 0.79&0.77& 0.75\\
 SVM&   0.76  &0.79 & 0.74   &0.76&0.76\\
 Random Forest&   0.80 &0.82 & 0.78 &0.80 & 0.80\\
 \textbf{XGBoost}& 0.80 &0.81&0.81&{0.81}& {0.80}\\
 \hline
 \multicolumn{6}{|c|}{\textbf{ADASYN}} \\
 \hline
 Logistic Regression   &0.70    &0.75& 0.71&0.73& 0.70\\
 SVM&   0.66  &0.72 & 0.63   &0.67&0.66\\
 \textbf{Random Forest}&   0.74 &0.78 & 0.75 &{0.77} & {0.74}\\
 XGBoost& 0.72 &0.75&0.74&0.75&0.71\\
 \hline
\end{tabular}
\end{table}

\begin{table}[!ht]
\caption{Pima Diabetes Data}
\label{PIMA}
\begin{tabular}{ |p{3cm}||p{2cm}|p{2cm}|p{2cm}|p{2cm}|p{2cm}|}
 \hline
 \multicolumn{6}{|c|}{\textbf{Original data set}} \\
 \hline
 Classifier & Accuracy&Precision&Recall&F1-score&AUC\\
 \hline
 Logistic Regression   &0.82    &0.76& 0.62&0.68&0.77\\
 SVM&   0.79  &0.70 & 0.55  &0.62 & 0.73\\
 Random Forest&   0.81  &0.71 & 0.64   &0.67 & 0.76\\
 \textbf{XGBoost}& 0.82 &0.70&0.70&{0.70}& {0.79}\\
 \hline
 \multicolumn{6}{|c|}{{SMOTE}} \\
 \hline
 Logistic Regression   &0.81    &0.78& 0.82&0.80&0.81\\
 \textbf{SVM}&   0.85  &0.83 & 0.87  &{0.85} & {0.86}\\
 \textbf{Random Forest}&   0.85  &0.83 & 0.87  &{0.85} & {0.86}\\
 XGBoost& 0.85 &0.82&0.88& 0.85&0.85\\
\hline
 \multicolumn{6}{|c|}{\textbf{ADASYN}} \\
 \hline
 Logistic Regression   &0.72    &0.73&0.71&0.72&0.72\\
 SVM&   0.82 &0.79 & 0.87   &0.83&0.82\\
 \textbf{Random Forest}&   0.85  &0.82 & 0.90  &{0.86} & {0.85}\\
 XGBoost& 0.82&0.78&0.89& 0.83&0.82\\
 
 \hline
\end{tabular}
\end{table}

\begin{figure}[H]
  \begin{minipage}[b]{0.49\textwidth}
    \centering
    \includegraphics[width=1\textwidth]{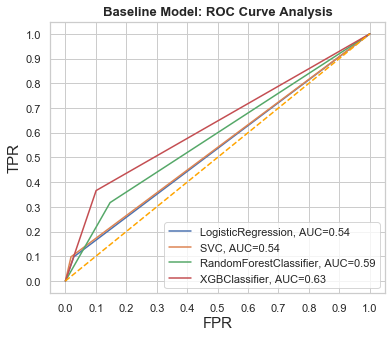}
  \end{minipage}
  \begin{minipage}[b]{0.49\textwidth}
    \centering
    \includegraphics[width=1\textwidth]{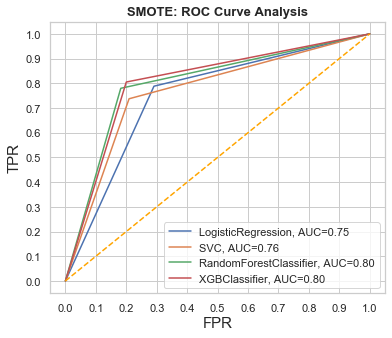}
  \end{minipage}
  \hspace{-0.5cm}
  \begin{minipage}[b]{0.49\textwidth}
   \centering
    \includegraphics[width=1\textwidth]{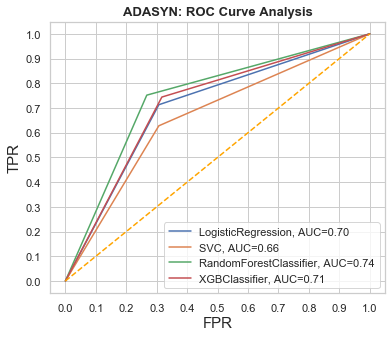}
  \end{minipage}
  \caption{Blood Transfusion Service Center Data}
  \label{BT}
\end{figure}

\begin{figure}[H]
  \begin{minipage}[b]{0.49\textwidth}
    \centering
    \includegraphics[width=1\textwidth]{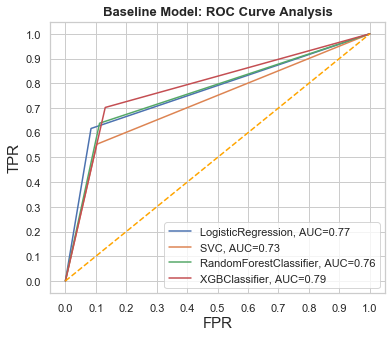}
  \end{minipage}
  \begin{minipage}[b]{0.49\textwidth}
    \centering
    \includegraphics[width=1\textwidth]{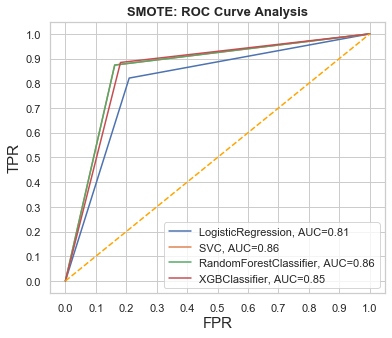}
  \end{minipage}
  \hspace{-0.5cm}
  \begin{minipage}[b]{0.49\textwidth}
   \centering
    \includegraphics[width=1\textwidth]{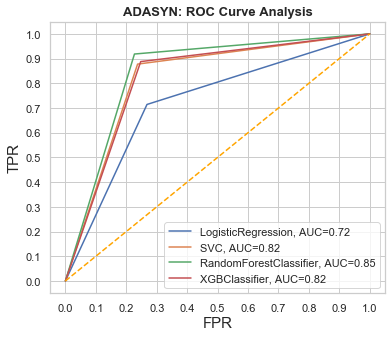}
  \end{minipage}
  \caption{Pima Diabetes Data}
  \label{Diab}
\end{figure}

\begin{figure}[H]
  \begin{minipage}[b]{0.49\textwidth}
    \centering
    \includegraphics[width=1\textwidth]{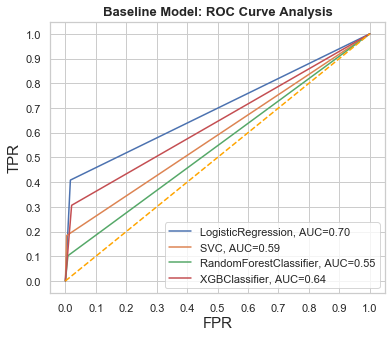}
 \end{minipage}
  \begin{minipage}[b]{0.49\textwidth}
    \centering
    \includegraphics[width=1\textwidth]{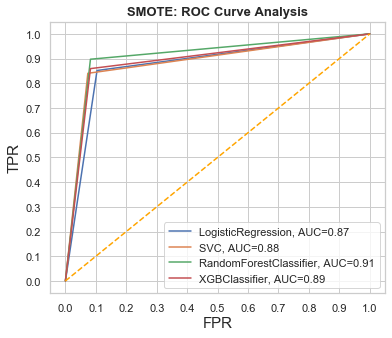}
  \end{minipage}
  \begin{minipage}[b]{0.49\textwidth}
   \centering
    \includegraphics[width=1\textwidth]{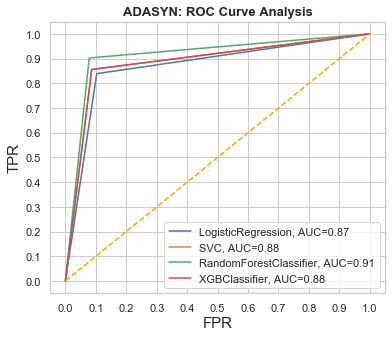}
  \end{minipage}
  \caption{ IBM HR Analytics Employee Attrition Data}
  \label{HR}
\end{figure}

\begin{table}[!ht]
\caption{IBM HR Analytics Employee Attrition Data} 
\label{IBM}
\begin{tabular}{ |p{3cm}||p{2cm}|p{2cm}|p{2cm}|p{2cm}|p{2cm}|}
 \hline
 \multicolumn{6}{|c|}{\textbf{Original Data Set}} \\
 \hline
 Classifier & Accuracy&Precision&Recall&F1-score&AUC\\
 \hline
 \textbf{Logistic Regression}   &0.89    &0.83& 0.41&{0.55}&{0.70}\\
 SVM&   0.86&0.90& 0.20  &0.31&0.59\\
 Random Forest&   0.84  &0.71& 0.10 &0.18 & 0.55\\
 XGBoost& 0.87 &0.75&0.31& 0.43&0.64\\
 \hline
 \multicolumn{6}{|c|}{\textbf{SMOTE}} \\
 \hline
 Logistic Regression   &0.87    &0.90& 0.85&0.88&0.87\\
 SVM&   0.88 &0.93 & 0.84   &0.88&0.88\\
\textbf{Random Forest}&   0.91  &0.93 & 0.90  &{0.91} & {0.91}\\
 XGBoost& 0.89 &0.92&0.86& 0.89&0.89\\
 \hline
 \multicolumn{6}{|c|}{\textbf{ADASYN}} \\
 \hline
 Logistic Regression   &0.87    &0.89& 0.84&0.86& 0.87\\
 SVM&   0.89  &0.91 & 0.86  &0.88&0.88\\
 \textbf{Random Forest}&   0.91 &0.92 & 0.90 {0.91} &{0.91}\\
 XGBoost& 0.89 &0.91&0.86&0.88&0.88\\
 \hline
\end{tabular}
\end{table}
\newpage
Figure \ref{BT}, \ref{Diab} \& \ref{HR} shows the ROC graph for the baseline models and the models after the application of the synthetic oversampling techniques. As seen in Figure \ref{BT}, \ref{Diab} \& \ref{HR}, there were improvement in the classifiers ability to distinguish between the minority and majority classes after applying SMOTE or ADASYN to the Imbalanced data. Random Forest and XGBoost when combined with SMOTE or ADASYN performed better than Logistic Regression and Support Vector Machine. As shown in Figure \ref{BT}, the combination of SMOTE and the classifiers outperformed ADASYN. Random Forest or XGBoost when combined with SMOTE yielded an AUC score of 0.80. That means that 80\% of the time the model is able to distinguish between a blood donor from a non-blood donor. With the Pima Diabetes Data, a combination of the SMOTE technique and the Random Forest classifier (Figure \ref{Diab}) gave the optimal AUC Score of 0.86 compared with the baseline model of 0.76. This implies that the model is able to distinguish between a patient who is diabetic and a patient who is not 86\% of the time as compared to 76\% for the baseline model. Random Forest classifier on the IBM HR Analytics Employee Attrition Data  reported an AUC score of 91\% after applying both SMOTE and ADASYN (Figure \ref{HR}).\par
In general, we observed in this paper that, the reporting measures scores were better after dealing with data Imbalance using SMOTE or ADASYN. However, choosing the best combination of sampling technique and classifier are essential to handling the problem of Imbalanced data in the best possible way.

\section{Conclusion}
In this paper, we focused on handling Imbalanced data  with particular focus on two re-sampling methodologies - SMOTE and ADASYN. We also focused on explaining the algorithms behind this techniques with computed examples. We discussed that, compared to SMOTE, ADASYN put more focus on the minority samples that are difficult to learn by generating more synthetic data points for these difficult and hard to learn minority sample class. We also discussed that while SMOTE considers a  uniform weight in generating new synthetic data for all minority points, ADASYN considers a density distribution in deciding the number of synthetic samples to be generated for a particular minority data point. We also from the experimental study realized how the reporting measure. accuracy could be misleading and proposed like many other researchers to rely on other metrics in conjunction with accuracy when reporting. While both SMOTE and ADASYN when applied to the Imbalanced data, resulted in an improvement in the reporting measures, SMOTE generally reported higher reporting measure scores than ADASYN. However, there are not enough evidence to generalize based on this paper that, SMOTE performs better than ADASYN in handling Imbalanced data. Factors not limited to the type of classifier, parameter tuning and the type of Imbalanced data can definitely affect the performances of SMOTE and ADASYN.\par
For future research, we will focus on how the combination of the oversampling techniques- SMOTE or ADASYN and classifiers perform on a test data that has different Imbalance sample class ratios from the train data set.  


\newpage

\end{document}